\documentclass[runningheads]{llncs}
\usepackage{graphicx}
\usepackage{comment}
\usepackage{amsmath,amssymb} 
\usepackage{color}
\usepackage{algorithm}
\usepackage{algorithmic}
\usepackage{indentfirst}
\usepackage{multirow}
\usepackage{array}
\usepackage{subfigure}
\usepackage[width=122mm,left=12mm,paperwidth=146mm,height=193mm,top=12mm,paperheight=217mm]{geometry}

\begin{document}
\pagestyle{headings}
\mainmatter
\def\ECCVSubNumber{3475}  

\title{Attribute Mix: Semantic Data Augmentation for Fine-grained Recognition} 

\titlerunning{Attribute Mix: Semantic Data Augmentation for Fine Grained Recognition}
%
\author{Hao Li\inst{1}\thanks{This work was done when the first author was an intern at Huawei Noah's Ark Lab.
} \and
Xiaopeng Zhang\inst{2}\and
Hongkai Xiong\inst{1}\and Qi Tian\inst{2}}
\authorrunning{Hao Li et al.}
%
\institute{Shanghai Jiaotong University \and
Huawei Noah's Ark Lab 
}

\maketitle
\vspace{-0.6cm}
\begin{abstract}
  Collecting fine-grained labels usually requires expert-level domain knowledge and is prohibitive to scale up. In this paper, we propose Attribute Mix, a data augmentation strategy at attribute level to expand the fine-grained samples. The principle lies in that attribute features are shared among fine-grained sub-categories, and can be seamlessly transferred among images. Toward this goal, we propose an automatic attribute mining approach to discover attributes that belong to the same super-category, and Attribute Mix is operated by mixing semantically meaningful attribute features from two images. Attribute Mix is a simple but effective data augmentation strategy that can significantly improve the recognition performance without increasing the inference budgets. Furthermore, since attributes can be shared among images from the same super-category, we further enrich the training samples with attribute level labels using images from the generic domain. Experiments on widely used fine-grained benchmarks demonstrate the effectiveness of our proposed method. 

\keywords{Fine-grained Recognition, Attribute Augmentation, Semi-supervised Learning}
\end{abstract}

\section{Introduction}

\noindent Fine-grained recognition aims at discriminating sub-categories that belong to the same general category, \emph{i.e.,} recognizing different kinds of birds~\cite{wah2011caltech,Berg2014BirdsnapLF}, dogs~\cite{KhoslaYaoJayadevaprakashFeiFei_FGVC2011}, and cars~\cite{KrauseStarkDengFei-Fei_3DRR2013} \emph{etc.}.  Different from general category recognition, fine-grained sub-categories often share the same parts (\emph{e.g.}, all birds should have wings, legs, heads, \emph{etc.}), and usually can only be distinguished by the subtle differences in texture and color properties of these parts (\emph{e.g.}, only the breast color counts when discriminating some similar birds). Although the advances of Convolutional Neural Networks (CNNs) have fueled remarkable progress for general image recognition \cite{hu2018squeeze,simonyan2014very,he2016deep,wang2017residual}, fine-grained recognition still remains to be challenging where discriminative details are too subtle to discern.
\begin{figure}[t]
  \centering
  \subfigure[Image $x_a$]{\includegraphics[width=0.18\textwidth]{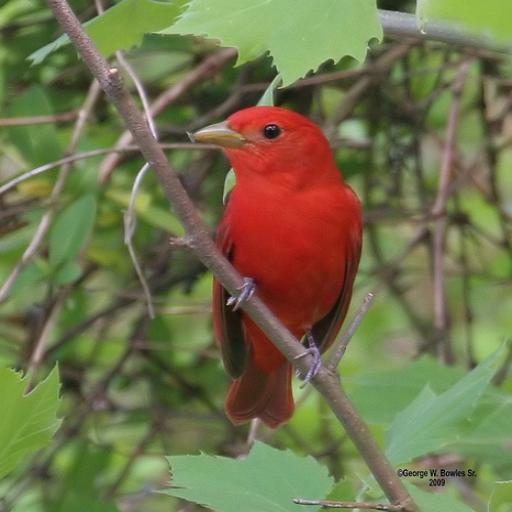}} \hspace{0.17cm}
  \subfigure[Image $x_b$]{\includegraphics[width=0.18\textwidth]{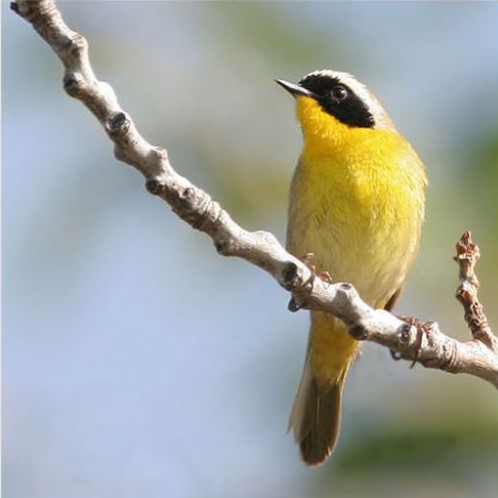}} \hspace{0.17cm}
  \subfigure[${M}_{a,1} \odot x_a$]{\includegraphics[width=0.18\textwidth]{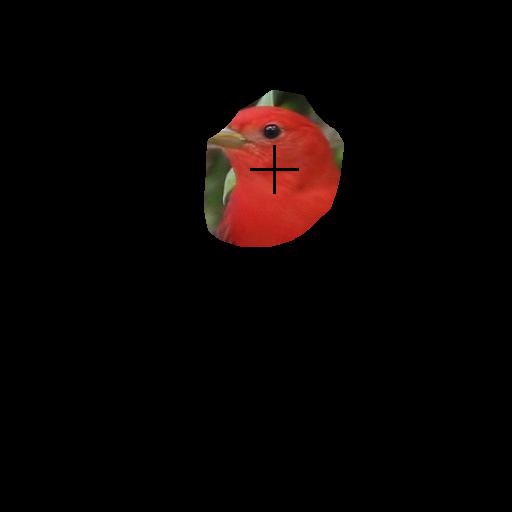}} \hspace{0.17cm}
  \subfigure[${M}_{b,1} \odot x_b$]{\includegraphics[width=0.18\textwidth]{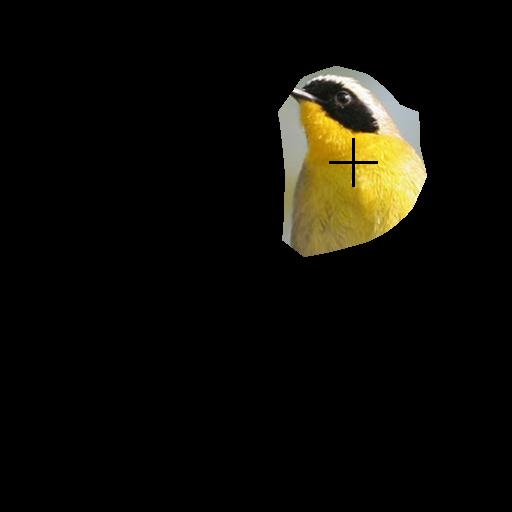}} \\
  \vspace{-0.1cm}
  \subfigure[Mixup]{\includegraphics[width=0.18\textwidth]{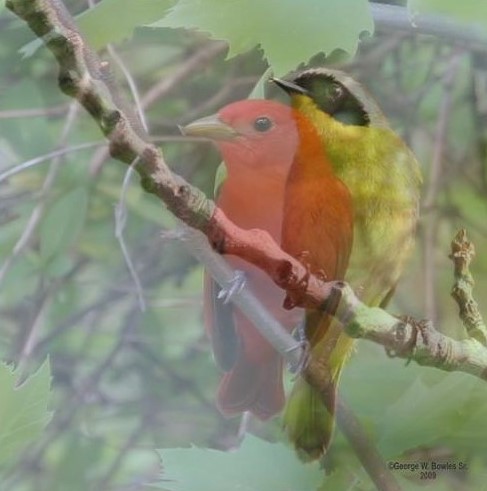}} \hspace{0.17cm}
  \subfigure[CutMix]{\includegraphics[width=0.18\textwidth]{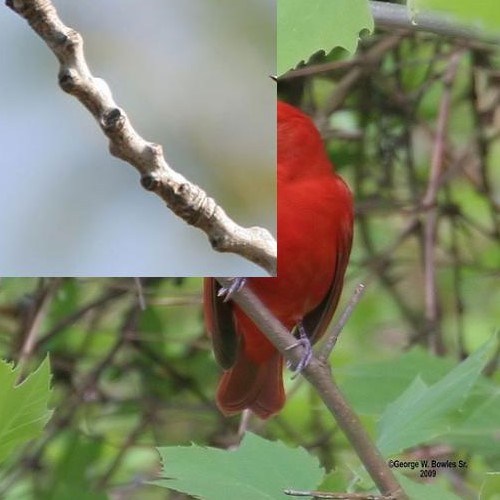}} \hspace{0.17cm}
  \subfigure[A-M]{\includegraphics[width=0.18\textwidth]{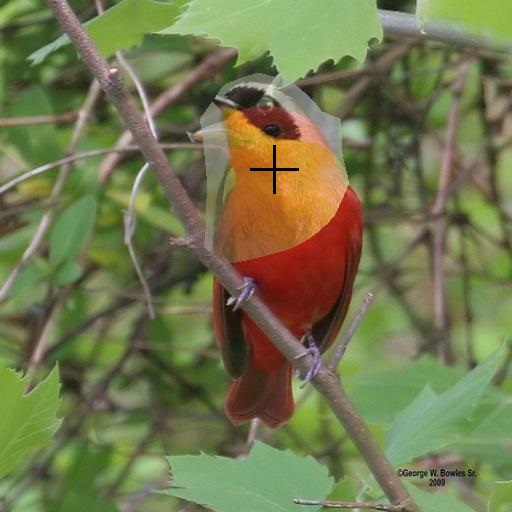}} \hspace{0.17cm}
  \subfigure[A-M]{\includegraphics[width=0.18\textwidth]{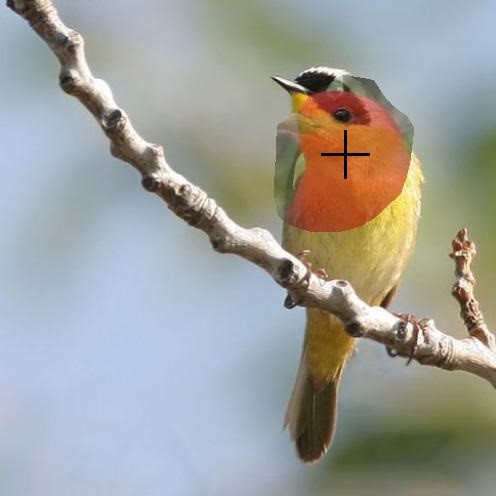}} \\
  \vspace{-0.2cm}
  \caption{Data augmentation via Mixup, CutMix and Attribute Mix. (a)(b) are samples $x_a$ and $x_b$ drawn at random from CUB-200-2011. (c)(d) are two mined discriminative attribute regions from $x_a$ and $x_b$, respectively. (e)(f) are virtual samples generated by Mixup and CutMix. (g)(h) are two virtual samples generated by Attribute Mix.}
  \label{Mix-result}
 \vspace{-0.3cm}
\end{figure}

Existing deep learning based fine-grained recognition approaches usually focus on developing better models for part localization and representation. Typical strategies include: 1) part based methods that first localize parts, crop and amplify the attended parts, and concatenate part features for recognition~\cite{lin2015deep,zhang2014part,branson2014bird,zhang2015fine}; 2) attention based methods that use visual attention mechanism to find the most discriminative regions of the fine-grained images~\cite{xiao2015application,fu2017look,zheng2017learning}; 3) feature based methods such as bilinear pooling~\cite{lin2015bilinear} or trilinear pooling~\cite{zheng2019looking} for better representation. However, we argue that for fine-grained recognition, the most critical challenge arises from the limited training samples, since collecting labels for fine-grained samples often requires expert-level domain knowledge, which is difficult to extend to large scale. As a result, existing deep models are easy to overfit to the small scale training data, and this is especially true when deploying with more complex modules.

In this paper, we promote fine-grained recognition via enriching the training samples at low cost, and propose a simple but effective data augmentation method to alleviate the overfitting and improve model generalization. Our method, termed as Attribute Mix, aims at enlarging the training data substantially via mixing semantically meaningful attribute features from two images. The motivation is that attribute level features are the key success factors to discriminate different sub-categories, and are transferable since all sub-categories share the same attributes. Toward this goal, we propose an automatic attribute learning approach to discover attribute features. This is achieved by training a multi-hot attribute level deep classification network through iteratively masking out the most discriminative parts, hoping that the network can focus on diverse parts of an object. The new generated images, accompanied with attribute labels mixed proportionally to the extent that two attributes fuse, can greatly enrich the training samples while maintaining the discriminative semantic meanings, and thus greatly alleviate the overfitting issue and are beneficial to improve model generalization.

Attribute Mix shares similarity with MixUp~\cite{zhang2017mixup} and CutMix~\cite{yun2019cutmix}, which all mix two samples by interpolating both images and labels. The difference is that both MixUp and CutMix randomly mix images or patches from two images, without considering their semantic meanings. As a result, it is usual for these two mixing operations to fuse images with non-discriminative regions, which in turn introduces noise and makes the model unstable for training. An example is shown in Fig.\ref{Mix-result}.  In comparison, Attribute Mix intentionally fuses two images at the attribute level, which results in more semantically meaningful images, and is helpful for improving model generalization.

Benefiting from the discovered attributes, we are able to introduce more training samples at attribute level with only generic labels. Here we denote the generic labels as single bit supervision indicating whether an object is from a general category, \emph{e.g.,} whether an object is a bird or not. We claim that the attribute features can be seamlessly transferred from the generic domain to the fine-grained domain without knowing the object's fine-grained sub-category labels. This is achieved by a standard semi-supervised learning strategy that mines samples at the attribute levels. The mined samples, intrinsically with mixed attribute labels, and we term this proposed method which mines attribute from the general domain as Attribute Mix+. Note that although fine-grained labels are difficult to obtain, it is much easier to obtain a general label of an object. In this way, we are able to conveniently scale the fine-grained training samples via mining attributes from the generic domain for better performance investigation.

Our proposed data augmentation strategy is a general framework at low cost, and can be combined with most state-of-the-art fine-grained recognition methods to further improve the performance. Experiments conducted on several widely used fine-grained benchmarks have demonstrated the effectiveness of our proposed method.  We hope that our research on attribute-based data augmentation could offer useful guidelines for fine-grained recognition.

To sum up, this paper makes the following contributions:

$\bullet$ We propose Attribute Mix, a data augmentation strategy to alleviate the overfitting for fine-grained recognition.

$\bullet$ We propose Attribute Mix+, which mines fine-grained samples at attribute level, and does not need to know the specific sub-category labels of the mined samples. Attribute Mix+ is able to scale up the fine-grained training for better performance investigation conveniently.

$\bullet$ We evaluate our methods on three challenging datasets (CUB-200-2011, FGVC-Aircraft, Standford Cars), and achieve superior performance over the state-of-the-art methods.

\section{Related Work}

\noindent We briefly review some works for fine-grained recognition, as well as some recent technologies in data augmentation, which are most related with our work.
\subsection{Fine-grained Recognition}
\noindent Fine-grained recognition has been studied for several years. Early works on fine-grained recognition focus on leveraging extra parts and bounding box annotations to localize the discriminative regions of an object~\cite{branson2014bird,zhang2014part,zhang2015fine,lin2015deep}.
Later, some weakly supervised localization methods ~\cite{bency2016weakly,singh2017hide,kim2017two,oquab2015object,zhou2016learning,selvaraju2017grad,zhang2018adversarial} are proposed to localize objects with only image level annotations.  In~\cite{zhou2016learning}, Zhou \textit{et al.} proposed to localize the objects by picking out the class-specific feature maps. In~\cite{zhang2018adversarial}, Zhang \textit{et al.} proposed to use adversarial training to locate the integral object and achieved superior localization results.

On the other hand, powerful features have been provided by better CNN networks.  Lin \textit{et al.}~\cite{lin2015bilinear} proposed a bilinear structure to compute the pairwise feature interactions by two independent CNNs. And it turns out that higher-order features interaction can make the features highly discriminative ~\cite{cui2017kernel}. To model the subtle differences between two fine-grained sub-categories, attention mechanism~\cite{xiao2015application,zheng2017learning,fu2017look} and metric learning~\cite{schroff2015facenet,cui2016fine} are often used.  Besides, Zhang \textit{et al.}~\cite{zhang2016picking} proposed to unify CNN with spatially weighted representation by Fisher Vectors, which achieves high performances on CUB-200-2011. Although promising performance has been achieved, these methods are all at the expense of higher computational cost, and are prohibitive to deploy on low-end devices.

Few works rely on external data to help facilitate recognition. Cui \textit{et al.}~\cite{cui2018large} proposed to use Earth Mover's Distance to estimate the domain similarity, and transfer the knowledge from the source domain which is similar to the target domain. In~\cite{krause2016unreasonable}, Krause \textit{et al.} collected millions of images with tags from Web and utilized the Web data by transfer learning. However, both methods make use of a large amount of class-specific labels for transfer learning. In this paper, we demonstrate that for fine-grained recognition, transferring attribute features is a powerful proxy to improve the recognition accuracy at low cost, without knowing the class-specific labels of the source domain.

\subsection{Data Augmentation}
\noindent Data augmentation can greatly alleviate overfitting in training deep networks. Simple transformations such as horizontal flipping, color space augmentations, and random cropping are widely used in recognition tasks to improve generalization. Recently, automatic data augmentation techniques, \emph{e.g.}, AutoAugment~\cite{cubuk2019autoaugment}, are proposed to search for a better augmentation strategy among a large pool of candidates. Differently, Mixup~\cite{zhang2017mixup} combined two samples linearly in pixel level, where the target of the synthetic image was the linear combination of one-hot label. Though not meaningful for human perception, Mixup has been demonstrated surprisingly effective for the recognition task. Following Mixup, there are a few variants \cite{verma2018manifold,guo2019mixup} as well as a recent effort named Cutmix \cite{yun2019cutmix}, which combined Mixup and Cutout \cite{devries2017improved} by cutting and pasting patches. However, all these methods would inevitably introduce unreasonable noise due to augmentation operations on random patches of an image, without considering their semantic meanings. Our method is similar to these methods in mixing two samples for data augmentation. Differently, the mixing operation is only performed around those semantic meaningful regions, which enables the model for more stable training and is beneficial for generalization.

\begin{figure*}[t]
  \begin{center}
        \includegraphics[width=0.98\linewidth,height=5cm]{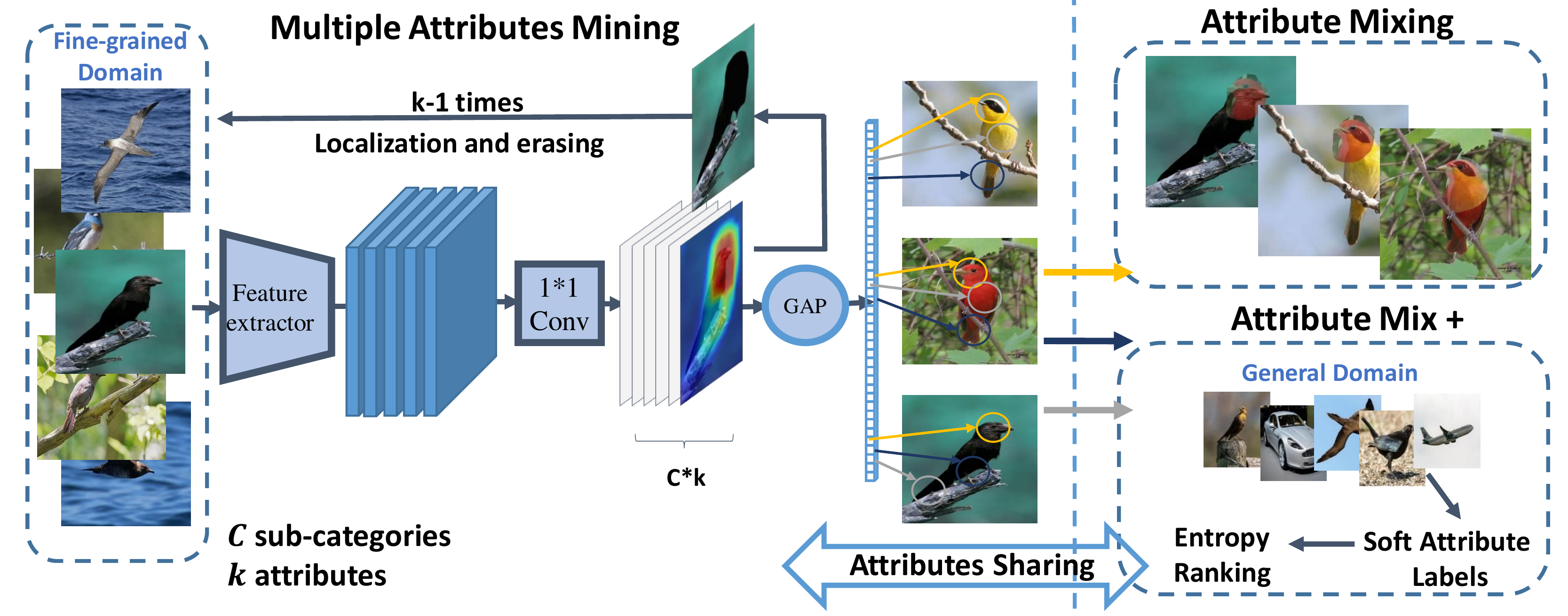}
  \end{center}
  \vspace{-0.3cm}
     \caption{The framework of the proposed Attribute Mix data augmentation. First, an automatic attribute learning approach is proposed to mine attribute features from a small scale fine-grained samples, implemented by training a multi-hot attribute level deep classification network through iteratively masking out the most discriminative parts. These attributes information can be used to generate new samples by Attribute Mix and shared between images from the same generic domain at attribute level.}
  \label{net_structure}
  \vspace{-0.2cm}
\end{figure*}

\section{Approach}
\noindent In this section, we describe our proposed attribute-based data augmentation strategy in detail. As shown in Fig.~\ref{net_structure}, the core ingredients of the proposed method consist of three modules: 1) Automatic attribute mining, which aims at discovering attributes with only image level labels. 2) Attribute Mix data augmentation, which mixes attribute features from any two images for new image generation. 3) Attribute Mix+, which enriches training samples via mining images from the same generic domain at attribute level.

\subsection{Attribute Mining}
\noindent The attribute level features are the core of the following data augmentation operation. We first elaborate how to obtain attribute level features with only image level labels. Denote $\{x,y\}, y\in \{0,1\}^C$ as a training image and its corresponding one-hot label with $y_c=1$, where $C$ is the number of fine-grained sub-categories. Without loss of generality, assuming that all fine-grained sub-categories share $k$ attributes, we simply convert the $C$ class level labels to more detailed, $k\times C$ attribute level labels for attribute discovery, as shown in Fig.~\ref{convert_label}. Specifically, the one-hot label $y$ of image $x$ is extended to multi-hot label $y^A$, while $\sum_{i=1}^C y_i =1$ and $\sum_{i=1}^{kC} y^A_i =k$, with each none zero hot regarded as one attribute corresponding to a specific sub-category. As shown in Fig.~\ref{net_structure}, for a typical CNN, we simply remove all the fully connected layers, and add a $1\times1$ convolutional layer to produce feature maps $f \in \mathbb{R}^{h \times w \times kC}$ with $kC$ channels, here every adjacent $k$ channels correspond to $k$ attributes for a certain sub-category. These feature maps are fed into a GAP (Global Average Pooling) layer to aggregate attribute level features for classification. The multiple attribute mining is proceeded as follows:

\begin{figure}[t]
  \centering
  \includegraphics[width=1\linewidth,height=4.2cm]{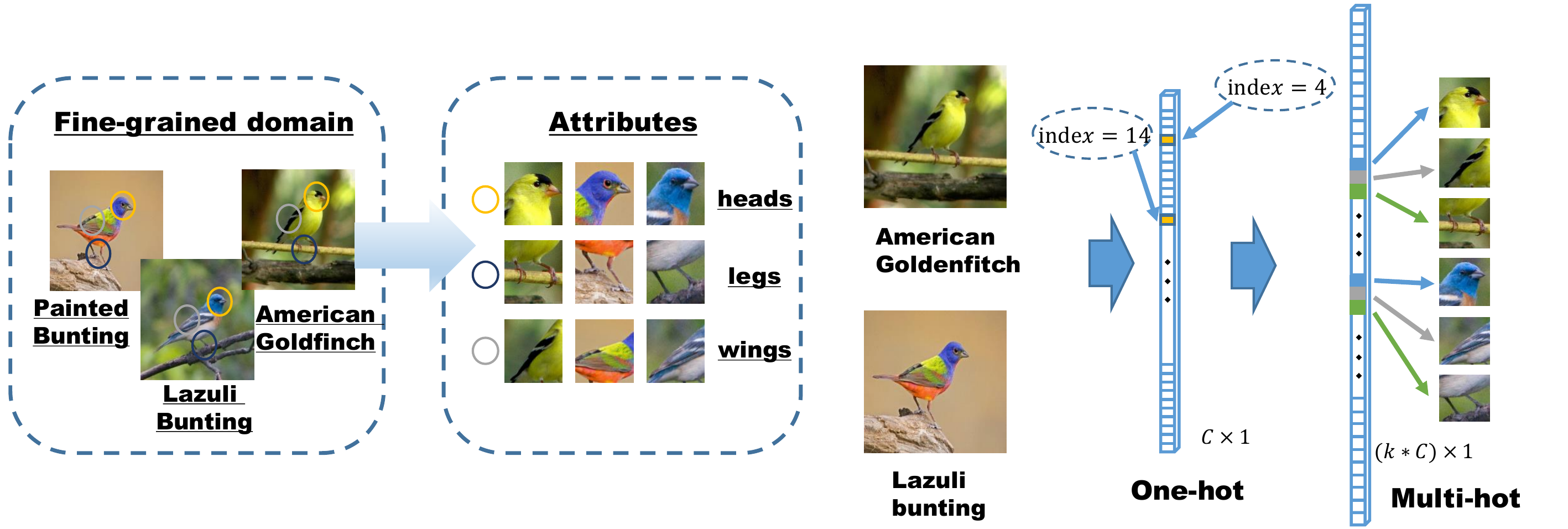}
  \vspace{-0.2cm}
  \caption{All fine-grained sub-categories from the same generic domain often share the same parts. The one-hot label is extended to multi-hot label from $C$ dimension to $k*C$ dimension, where each sub-category is endowed with $k$ attributes. For simplicity, we order the attributes belonging to the same sub-category at adjacent locations.}\label{convert_label}
  \vspace{-0.2cm}
\end{figure}

\begin{enumerate}
  \vspace{-0.1cm}
  \setlength{\itemsep}{1pt}
  \setlength{\parsep}{1pt}
  \setlength{\parskip}{1pt}
 \item Training multi-hot attribute level classification network with original images and attribute level multi-hot labels $y^A_{ck:(c+1)k-1}=1$.
 \item For an image with label $y^A_c$, picking out the corresponding feature map at the $kc$th channel $f(:,:, kc)$, generating the attention map according to the activations and upsampling to the original image size, thus we obtain the most discriminative region mask $M_{c,1}$ of an image.
 \item $M_{c,1}$ is used to erase the original image $x$ to get erased image $x^{M_{c,1}}$, and the corresponding multi-hot label, $y^A_{ck}$ changes from $1$ to $0$.
 \item Using $x^{M_{c,1}}$ as a new training sample, and do the above three steps to obtain masks $M_{c,i}$, i=2,...,k for all remained attributes.
\end{enumerate}

\begin{figure}[t]
  \centering
  \subfigure[Raw image]{\includegraphics[width=0.18\textwidth]{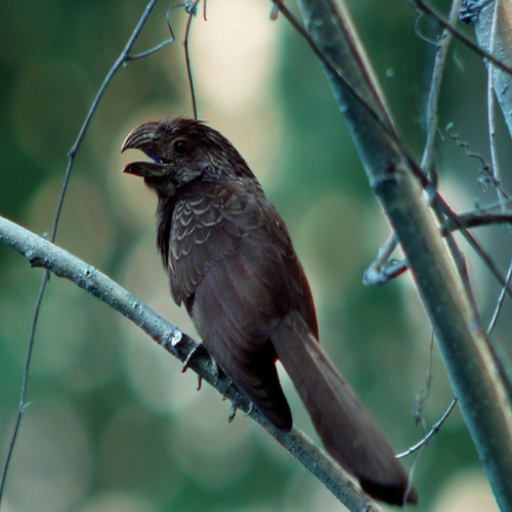}} \hspace{0.17cm}
  \subfigure[Part-1]{\includegraphics[width=0.18\textwidth]{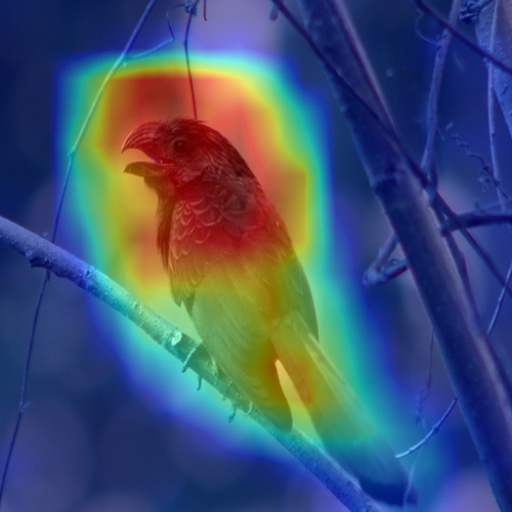}} \hspace{0.17cm}
  \subfigure[Part-2]{\includegraphics[width=0.18\textwidth]{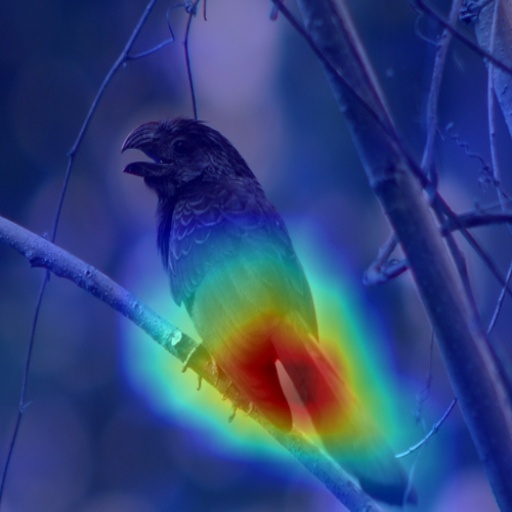}} \hspace{0.17cm}
  \subfigure[Part-3]{\includegraphics[width=0.18\textwidth]{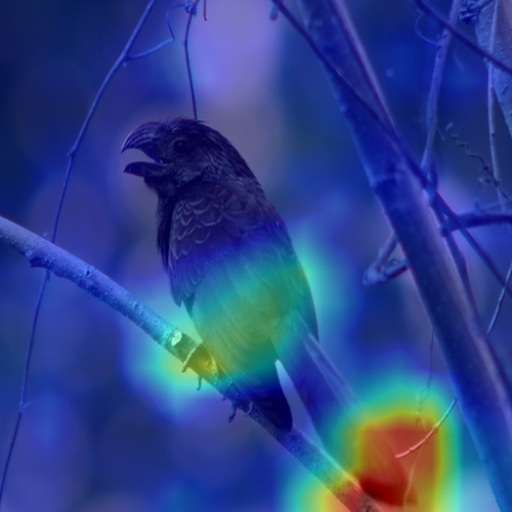}}  \\
  
  \subfigure[Raw image]{\includegraphics[width=0.18\textwidth]{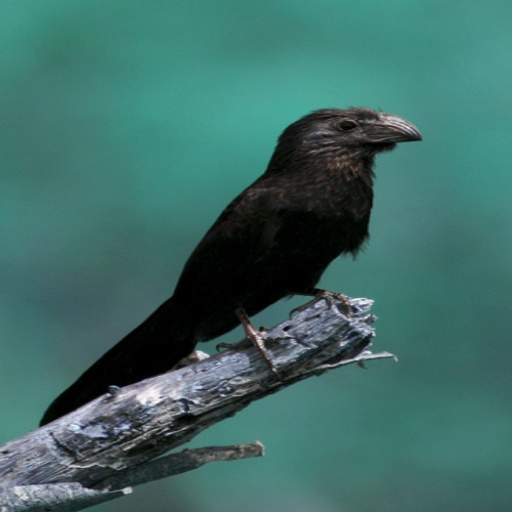}} \hspace{0.17cm}
  \subfigure[Part-1]{\includegraphics[width=0.18\textwidth]{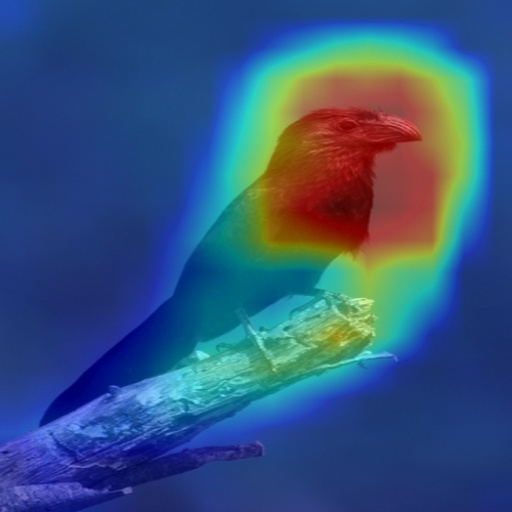}} \hspace{0.17cm}
  \subfigure[Part-2]{\includegraphics[width=0.18\textwidth]{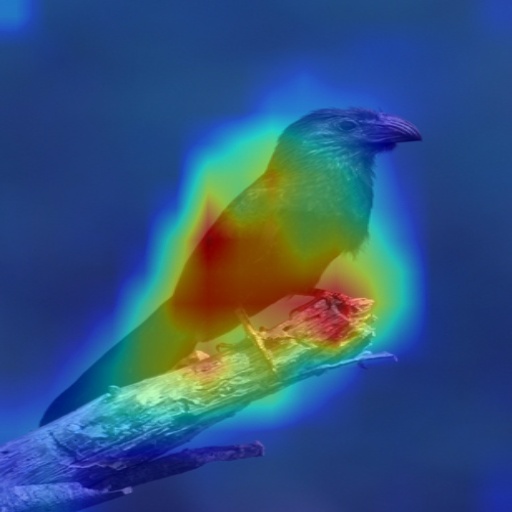}} \hspace{0.17cm}
  \subfigure[Part-3]{\includegraphics[width=0.18\textwidth]{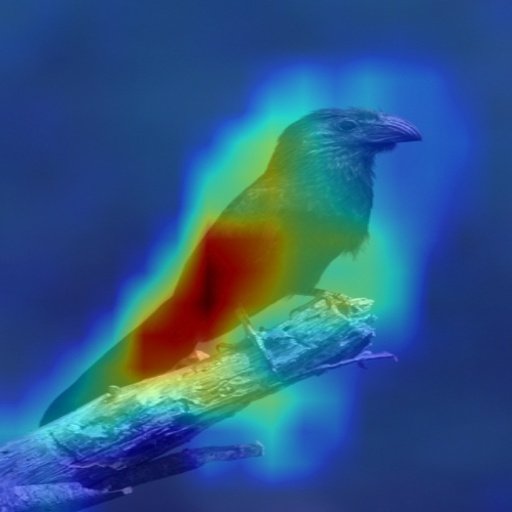}} \\
  \caption{Illustration of multiple attribute mining. Example images show the mined attributes when $k = 3$. These attributes approximately correspond to birds' heads, wings and tails, respectively.}
  \label{part-heatmap}
\end{figure}

Following the above procedures, we are able to train an attribute level classification network automatically and obtain a series of masks $M_c$, which correspond to different attributes of an object. Fig. \ref{part-heatmap} shows an example of what the Attribute Mining procedure learns. It can be shown that these attributes coarsely correspond to different parts of an object, \emph{e.g.}, the birds' heads, wings, and tails.

\subsection{Attribute Mix}
\noindent After obtaining the attribute regions, we introduce a simple data augmentation strategy on attribute level to facilitate fine-grained training. The proposed Attribute Mix operation constructs synthetic examples by intentionally mixing the corresponding attribute regions from any two images. Specifically, given two samples $(x_{a},y_{a}^A)$ and $(x_{b},y_{b}^A)$ with $x_a, x_b \in R^{W\times H\times 3}$, and random picked attribute masks $M_{a,k}$, $M_{b,k} \in \{0,1\}^{W\times H}$, the generated training sample $(\widetilde{x}, \widetilde{y})$ is obtained by:
\begin{equation}
  \begin{aligned}
    \widetilde{x} &= (\mathbf{1}-\widetilde{M}_{b,k})\odot x_a + \lambda \widetilde{M}_{b,k} \odot x_a + (1-\lambda) \widetilde{M}_{b,k} \odot x_b  \\
    \widetilde{y} &= \lambda y_a^A + (1-\lambda) y_b^A, \\
  \end{aligned}
  \label{A_fc}
\end{equation}

where $\widetilde{M}_{b,k}$ is the transformed binary mask from $x_b$ to $x_a$, with the mask center-aligned with $M_{a,k}$, and denotes the region that needs to be mixed. $\mathbf{1}$ is a binary mask filled with ones, and $\odot $ is the element-wise multiplication operation. Like Mixup \cite{zhang2017mixup}, the combination ratio $\lambda$ between two regions is sampled from the beta distribution $Beta(\alpha,\alpha)$. In all our experiments, we set $\alpha=1$, meaning that $\lambda$ is sampled from the uniform distribution $(0,1)$. An illustration of the Attribute Mix operation is shown in Fig.\ref{Mix-result} , as well as some comparison results that generated by Mixup and CutMix operations. Compared with Mixup and CutMix which inevitably introduce some meaningless samples (\emph{e.g.} random patches, background noise), Attribute Mix only focuses on the foreground attribute regions and is more suitable for fine-grained categorization.

Though more semantically meaningful our Attribute Mix operation is, the generated virtual samples still suffer large domain gap with the original, natural images. As a result, memorizing these samples would deteriorates the model generalization. To address this issue, we introduce a time decay learning strategy to limit the probability of applying Attribute Mix operation. This is achieved by controlling the mixing ratio $\lambda$ in Eq. (\ref{A_fc}), \emph{i.e.,} we introduce a variable $\gamma(t)$, which increases from 0 to 1 as the training proceeds, and limit $\gamma(t) \leq \lambda \leq 1$. In this way, $\lambda$ is sampled from $Beta(\alpha, \alpha)$ distribution, and only when $\lambda$ is larger than $\gamma(t)$, the generated samples are used for training.  As the training process goes on, the mixed operations between two images decay, and finally degenerate to using the original images. In the experimental section, we will validate its effectiveness in improving model generalization.

\subsection{Attribute Mix$+$}
\label{sec_attribute_transfer}
\noindent In principle, the attribute features are shared among images from the same super-category, regardless of their specific sub-category labels. This conveniently enables us to expand the training samples at attribute level to images from the generic domain. In this section, we enrich the training samples in another view, \emph{i.e.}, transfer attributes from images with only generic labels. This is achieved by generating soft, attribute level labels via a standard semi-supervised learning strategy over a large amount of images with low cost, general labels. Since attributes are shared among all the sub-categories, it is not necessary for images from the generic domain that belong to the same sub-categories with the target domain.

Using the model trained with Attribute Mix strategy, we conduct inference over the images from the generic domain, and produce attribute level probabilities by using a softmax layer. Specifically, denoting the model output as $z \in R^{kC}$, we reshape the output to $\widetilde{z} \in R^{k \times C}$, where each row corresponds to the same attribute among different sub-categories. The softmax operation is conducted over the row dimension to obtain the probability $p \in R^{k \times C}$:
\begin{equation}
  \label{softmax}
  p_{i,j} = \frac{exp(\widetilde{z}_{i,j}/T))}{\sum_{c=1}^C exp(\widetilde{z}_{i,c}/T))},
\end{equation}

where $T$ is a temperature parameter that controls the smooth degree of the probability.

 \textbf{Entropy ranking}.
In many semi-supervised learning methods \cite{grandvalet2005semi,berthelot2019mixmatch,lee2013pseudo,miyato2018virtual}, it is a common underlying assumption that classifiers' decision boundary should not pass through high-density regions of the marginal data distribution. 
Similarly, in our experiment, the collected data with generic labels inevitably contain noise that would probably hurt the performance, and some of them are with too smooth soft-labels that can not provide any valuable attribute information. To address this issue, we propose an entropy ranking strategy to select images adaptively. Specifically, given an image $x_g$ from the generic domain with soft attribute label $p$, its entropy is calculated as follows:

\begin{equation}
  \label{entropy}
  H(x_g)= -\sum_{i=1}^{k}\sum_{c=1}^{C}p_{i,c}\mathrm{log}_2 p_{i,c},
\end{equation}
where $p_{i,c}$ denotes the probability that image $x_g$ contains attribute $i$ of fine-grained sub-category $c$. $H(x_g)$ is large if the attribute distribution is smooth, and reaches its maximum when all attributes obtain the same probability, which we think carry no valuable information for fine-grained training.  Based on this property, we set the criteria that samples with entropy $H(x_g)$ that higher than threshold $\tau$ to be filtered out.  In the ablation study, we will validate its effectiveness for achieving stable results, especially when noisy images exist.

\section{Experiments}

\subsection{Datasets and Implementation Details}
\textbf{Datasets.}
The empirical evaluation is performed on three widely used fine-grained benchmarks: Caltech-USCD Birds-200-2011~\cite{wah2011caltech}, Standford Cars~\cite{KrauseStarkDengFei-Fei_3DRR2013}, and FGVC-Aircraft~\cite{maji2013fine}, which belong to three generic domains: birds, cars and aircraft, respectively. Each dataset is endowed with specific statistic properties, which are crucial for recognition performance. CUB-200-2011 is the most widely used fine-grained dataset, which contains 11,788 images spanning 200 sub-species. Standford Cars consists of 16,185 images with 196 classes, which are produced by different manufacturers, while FGVC-Aircraft consists of 10,000 images with 100 species. In current view, these are all small scale datasets. We use the default training/test split for experiments, which gives us around 30 training examples per class for birds, 40 training examples per category for cars, and approximately 66 training examples per category for aircraft.

\textbf{Implementation details.}
In our implementation, all experiments are based on ResNet-50~\cite{he2016deep} backbone. We first construct our baseline model over three datasets for following comparisons.
During network training, the input images are randomly cropped \textbf{(scale $s \sim U(0.15,1)$)} to $448\times448$ pixels after being resized to $512\times512$ pixels and randomly flipped. For Standford Cars, color jittering is used for extra data augmentation. We train the models for $80$ epochs, using Stochastic Gradient Descent (SGD) with the momentum of 0.9, weight decay of 0.0001.  The learning rate is set to $0.001$, which decays by a factor of $10$ every $30$ epochs.   During inference, the original image is \textbf{center cropped} and resized to $448 \times 448$. Benefiting from the powerful features, our baseline models have achieved considerable pleasing performance over these datasets. In the following, we will validate the effectiveness of the proposed method, even over such high baselines. 

For multi-hot attribute level classification, a standard binary cross-entropy loss is used. We increase the training epochs to $300$ after introducing Attribute Mixed samples, since it needs more rounds to converge for these augmented data. When mining attributes from the generic domain, we use the model pretrained with Attribute Mix to inference over data from the generic domains to produce soft attribute level labels. During inference, the multi-hot attribute level classification network outputs the predicted scores of $k*C$ attributes, and we simply combine the predicted scores for $k$ attributes of each sub-category.

\begin{table}[t]
\centering
\fontsize{9pt}{13pt}\selectfont
\setlength\tabcolsep{8pt}
  \caption{Performances comparison for different $\alpha$ on Attributes Mix}
  \label{k_table}
  \begin{tabular}{p{2cm}<{\centering}|p{3cm}<{\centering}|p{0.8cm}<{\centering}}
    \hline
    Dataset  &   Hyperparameter $\alpha$   & Acc. \\ \hline
    \multirow{4}{*}{ CUB-200-2011} & \textbf{$0.25$} & 87.9 \\
    & \textbf{$0.5$} & 88.2  \\
    & \textbf{$1$} & \textbf{88.4}  \\
    & \textbf{$2$} & 87.8  \\
    \hline

  \end{tabular}    
  \end{table}

\subsection{Ablation Study}
\noindent  In this section, we investigate some parameters which are important for recognition performance. Unless otherwise specified, all experiments are conducted on CUB-200-2011.

\begin{table}[t]
\centering
\fontsize{9pt}{13pt}\selectfont
\setlength\tabcolsep{8pt}
  \caption{Performances comparison for different $k$ on Attributes Mix}
  \label{alphatable}
  \begin{tabular}{p{2cm}<{\centering}|p{3cm}<{\centering}|p{0.8cm}<{\centering}}
    \hline
    Dataset  &   Number of attributes  $k$   & Acc. \\ \hline
    \multirow{4}{*}{ CUB-200-2011} & $1$ & 85.3 \\ 
    & \textbf{$2$} & 87.9 \\
    & \textbf{$3$} & \textbf{88.4}  \\
    & $4$ &  88.2 \\
    \hline

  \end{tabular}    
  \end{table}

\begin{table}[t]
\centering

\fontsize{9pt}{13pt}\selectfont                         
\setlength\tabcolsep{8pt}
  \caption{Impact of adaptive Attribute Mix on CUB-200-2011 top-1 Acc.}
  \label{decay_table}
  \begin{tabular}{c|c|c}
    \hline
    Dataset & Adaptive & Acc.\\ \hline
    \multirow{2}{*}{ CUB-200-2011} & \checkmark & \textbf{88.4}  \\
     & & $87.8$ \\
    \hline

  \end{tabular}

  \end{table}
  
\begin{table}[t]
\centering
\fontsize{9pt}{13pt}\selectfont
\setlength\tabcolsep{8pt}
 \caption{Comparison between attribute level and image-level sample mining.}
 \vspace{-0.2cm}
 \label{img_vs_attribute}
 \begin{tabular}{p{2.3cm}<{\centering}|p{2.3cm}<{\centering}|p{1.5cm}<{\centering}}
   \hline
   Dataset & Methods & Acc.\\ \hline
   \multirow{2}{*}{ CUB-200-2011} & image-level  & 88.0  \\
    & attribute level  & \textbf{89.2}  \\
   \hline

 \end{tabular}
 \end{table}

\begin{table}[t]
\centering

\fontsize{9pt}{13pt}\selectfont                         
\setlength\tabcolsep{8pt}
  \caption{Impact of threshold in entropy ranking.}
  \label{tau}
  \begin{tabular}{c|c|c}
    \hline
    Dataset & $\tau$ & Acc.\\ \hline
    \multirow{5}{*}{ CUB-200-2011} & 1.0 & $89.0$  \\
     & 1.5 & \textbf{$89.4$} \\
     & 2.0 & $89.3$ \\
     & 2.5 & $89.1$ \\
     & 3.0 & $88.8$ \\
    \hline

  \end{tabular}
  
  \end{table}

\textbf{Effects of number of attributes $k$.}
Here we inspect the recognition performance w.r.t. the number of attributes $k$ during Attribute Mixing. The performances for different choices of $k$ are shown in Table~\ref{k_table}. If we set $k$ too small, the model cannot mine adequate attributes of an object and the improvement is marginal. Specifically, $k=1$ denotes the baseline without multiple attribute mixing. While for larger $k$, the model is at the risk of including background clutters, and making the optimization difficult.

\textbf{Impact of hyperparameter $\alpha$.}
The hyperparameter $\alpha$ in Eq. (\ref{A_fc}) plays an important role during mixing, which controls the strength of interpolation between attributes from two training samples.  Here we try different choices with $\alpha \in \{0.25,0.5,1,2,4\}$.  The performances of different $\alpha$ are shown in the Table~\ref{alphatable}, and the best performance can be achieved when $\alpha$ is set to $1$. 

\textbf{Effects of adaptive Attribute Mix.}
We introduce the time decay strategy to alleviate the overfitting over the augmented samples. The probability of applying Attribute Mix decays from $1$ to $0$ as the $cosine$ curve. The comparison results on CUB-200-2011 are shown in Table~\ref{decay_table}. It shows that Attribute Mix with the time decay learning strategy achieves higher accuracy of $88.4\%$, which surpasses Attribute Mix without that strategy by $0.6$ points.

\textbf{Comparison with image level image mining.}
In Section~\ref{sec_attribute_transfer}, the images from the generic domain are used to mine attribute level features. In order to validate the advantages of attribute level features, we compare with the traditional, semi-supervised learning using only image level labels. Specifically, the image level pseudo labels directly leverage the information from unlabeled data $\mathcal{D}_{UL}$ without considering the attribute information. For fair comparison, both methods use ResNet-101 model and the Attribute Mix is not applied during training. We use the model's prediction on attribute level and image level separately to generate the soft-labels over those images from the generic domain. As shown in Table~\ref{img_vs_attribute}, our proposed attributes level features achieve much higher accuracy of $89.6\%$, which surpasses the image level result $88.3\%$ by $1.3$ points. It is not a small gain considering the high baseline we used and the difficulty of CUB-200-2011 dataset.




\textbf{Effects of $\tau$ for entropy ranking.}
When leverage the data from generic domain, we introduce the entropy ranking to select samples that share attributes contributing most for the fine-grained recognition.
The entropy ranking mechanism investigates the correlations between the mined images from generic domain and fine-grained domain, and is robust to noisy images that probably do not belong to the same generic domain. In order to inspect the effectiveness of entropy ranking at length, we intentionally introduce some extra images with labels different from the generic labels, and test the performance of our method under such a situation.
Specifically, we choose external dataset PASCAL VOC 2007~\cite{everingham2007pascal} and 2012 \cite{pascal-voc-2012} as noisy images, which both contain 20 object classes. This is a dataset with multi-label images, and the number of samples that include birds is 1,377, only a small ratio (around $6.4\%$) over the whole dataset. Overall, this dataset can be treated as adding noisy images (around $16.3\%$) to the generic domain dataset. We evaluate our method using these noisy data, and the results with different thresholds $\tau$ are shown in the T
able~\ref{tau}. When the threshold $\tau$ is set to $1.5$, entropy ranking mechanism can filter out most of the noisy samples in VOC $07+12$, only $1,833$ samples reserved.  We claim that the advantages of ranking mechanism are obvious.

\begin{table}[t]
\centering

\caption{Comparisions with state-of-the-art methods on CUB-200-2011, Standford Cars and FGVC-Aircrafts}
  \label{result_cub}
\fontsize{8.7pt}{12.5pt}\selectfont
\setlength\tabcolsep{4pt}
  \begin{tabular}{l|l|ccc}
  \hline
  Methods & Backbone & CUB & Cars & Aircrafts \\
  \hline
  Baseline& R-50 $^*$ & 85.3 & 91.7 & 88.5 \\
  MAMC~\cite{sun2018multi}& R-50 &86.2 & 92.8 & -   \\

  DBTNet~\cite{zheng2019learning}&R-50 & 87.5 & 94.1 & 91.2  \\
  S3N ~\cite{ding2019selective}&R-50 & 88.5 & 94.7 & 92.8 \\
  MGE-CNN ~\cite{zhang2019learning} & R-50 & 88.5 & 93.9 & - \\

  Mixup~\cite{zhang2017mixup}&R-50 $^*$ & 85.9 & 92.3 & 89.2  \\
  CutMix~\cite{yun2019cutmix}& R-50 $^*$ & 86.2 & 92.3 & 88.7 \\

  \hline
  Attribute Mix &R-50 $^*$ & \textbf{88.4} & \textbf{94.9} & \textbf{92.0}  \\
  Attribute Mix (MGE-CNN) &R-50 & \textbf{89.3} & \textbf{95.0} & \textbf{92.9}  \\
  Attribute Mix (S3N) &R-50 & \textbf{89.2} & \textbf{95.2} & \textbf{93.4}  \\
  \hline
  \multicolumn{5}{l}{$*$ denotes that using ResNet50 as backbone without any changes.}
  \end{tabular}

  \vspace{-0.2cm}
\end{table}

\subsection{Comparisons with State-of-the-arts}
\noindent We now move on to compare our proposed method with state-of-the-art works on above mentioned fine-grained datasets. In Table~\ref{result_cub}, we show the comparison results on CUB-200-2011, Standford Cars and FGVC-Aircraft.  For fair comparison, we choose recent works which use similar backbone with us. MAMC~\cite{sun2018multi} introduces complex attention modules to model the subtle visual differences. In \cite{zheng2019learning}, And bilinear interaction with high computational complexity is used in~\cite{zheng2019learning} to learn fine-grained image representations. S3N~\cite{ding2019selective} and MGE-CNN~\cite{zhang2019learning} use multiple ResNet-50 as multiple branches, which greatly increases the complexity.  As for our method, our proposed Attribute Mix achieves a superior accuracy of $88.4\%$, $94.9\%$ and $92.0\%$ on CUB-200-2011, Standford Cars and FGVC-Aircraft without any complicated modules.

Compared with the other data-mixing augmentation methods, Attribute Mix outperforms the Mixup and CutMix at least $2.2$ points, which further demonstrates that Attribute Mix is more suitable for fine-grained categorization.  In order  to validate  the  generality  of  our  method,  we  also  incorporate Attribute Mix into the advanced methods S3N and MGE-CNN, following all the parameters setting in \cite{ding2019selective}\cite{zhang2019learning}, and promote the accuracy to $89.2\%$, $95.2\%$ and $93.4\%$ with S3N, $89.3\%$, $95.0\%$ and $92.9\%$ with MGE-CNN.  It can be seen that Attribute Mix achieves the state-of-the-art performances on all these fine-grained datasets without any complex changes on the baseline model, and it can be also easily combined with the other SOTA methods to further improve the performance.

\section{Conclusion}
\noindent This paper presented a general data augmentation framework for fine-grained recognition. Our proposed method, named Attribute Mix, conducts data augmentation via mixing two images at attribute level, and can greatly improve the performance without increasing the inference budgets. Furthermore, based on the principle that the attribute level features can be seamlessly transferred from the generic domain to the fine-grained domain regardless of their specific labels, we enrich the training samples with attribute level labels using images from the generic domain with low labelling cost, and further boost the performance. Our proposed method is a general framework for data augmentation at low cost, and can be combined with most state-of-the-art fine-grained recognition methods to further improve the performance. Experiments conducted on several widely used fine-grained benchmarks have demonstrated the effectiveness of our proposed method.

%
%
\bibliographystyle{splncs04}
\bibliography{egbib}
\end{document}